\documentclass[10pt,twocolumn,letterpaper]{article}
\usepackage{cvpr}
\usepackage{times}
\usepackage{epsfig}
\usepackage{graphicx}
\usepackage{amsmath}
\usepackage{amssymb}
\usepackage{algorithm}
\usepackage{algorithmic}
\usepackage{setspace}
\usepackage{hyperref}
\usepackage[utf8]{inputenc}
\usepackage{authblk}

\cvprfinalcopy 

\def\cvprPaperID{71} 

\begin{document}


\title{Describe and Attend to Track: Learning Natural Language guided Structural Representation and Visual Attention for Object Tracking}
\author[1]{Xiao Wang}
\author[1]{Chenglong Li}
\author[1]{Rui Yang}
\author[2, 3]{Tianzhu Zhang}
\author[1]{Jin Tang \thanks{Professor Jin Tang is the corresponding author. \{wangxiaocvpr, lcl1314, yangruiahu\}@foxmail.com, \{luobin, tj\}@ahu.edu.cn, tzzhang@nlpr.ia.ac.cn}}
\author[1]{Bin Luo}
\affil[1]{School of Computer Science and Technology, Anhui University, Hefei, 230601, China}
\affil[2]{National Laboratory of Pattern Recognition, Institute of Automation, CAS}
\affil[3]{University of Chinese Academy of Sciences}

\maketitle

\begin{abstract}
The tracking-by-detection framework requires a set of positive and negative training samples to learn robust tracking models for precise localization of target objects. However, existing tracking models mostly treat different samples independently while ignores the relationship information among them. In this paper, we propose a novel structure-aware deep neural network to overcome such limitations. In particular, we construct a graph to represent the pairwise relationships among training samples, and additionally take the natural language as the supervised information to learn both feature representations and classifiers robustly. To refine the states of the target and re-track the target when it is back to view from heavy occlusion and out of view, we elaborately design a novel subnetwork to learn the target-driven visual attentions from the guidance of both visual and natural language cues. Extensive experiments on five tracking benchmark datasets validated the effectiveness of our proposed method.
\end{abstract}

\section{Introduction}

\begin{figure*}[t]
\center
\includegraphics[width=7in]{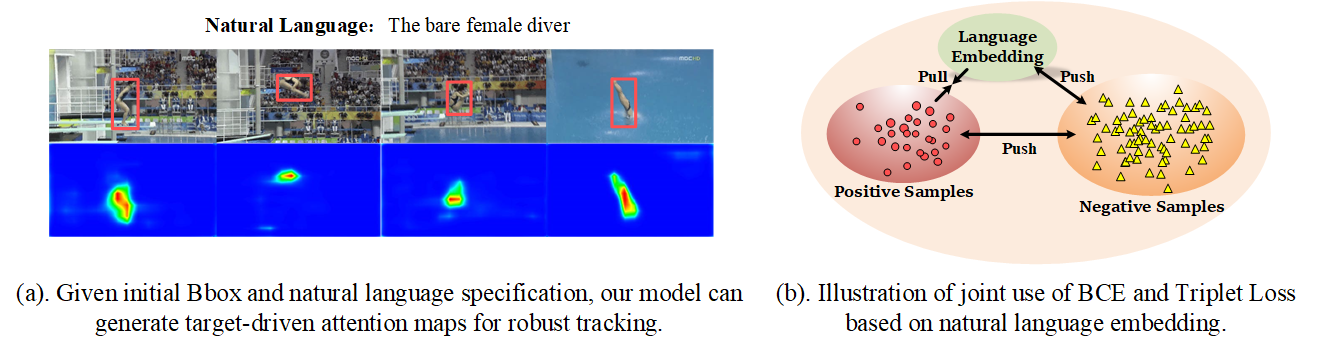}
\caption{The definition of tracking by natural language specification (left sub-figure) and loss functions we used for the optimization of our tracker (right sub-figure). }
\label{differencewithMDNet}
\end{figure*}

As a classical and challenge task in computer vision, visual tracking has been widely used in various applications, such as intelligent surveillance and automatic driving. Although appealing results have been achieved, visual tracking is still challenging partly due to the existences of the extreme factors including heavy occlusion, abruptly changing, large deformation and out of view.

Most successful visual trackers follow the tracking-by-detection framework, in which a set of positive and negative samples are used to train the parameters of classifiers and deep representations. However, existing tracking models mostly treat different training samples independently while ignores the relationship information among them, which is crucial to the robustness of feature representation and classifier learning. For instance, when we attempt to estimate the response score of one sample, existing approaches only consider the relationship between positive and negative samples. Other relations among different sample pairs are ignored. As a result, some hard positive or negative samples are difficult to obtain proper response score since only limited relationship information among samples is utilized for the overall estimation.

Recently, natural language is introduced in visual tracking, which achieves improved tracking performance~\cite{Li2017Tracking}. For example, Li \emph{et al.}~\cite{Li2017Tracking} propose three models including natural language only, visual target specification based on language, and leveraging their joint capacity, to help visual trackers against model drift. However, they use the Recurrent Neural Network (RNN) model to encode the input sentences to generate a dynamic filter, in which the RNN module would increase heavy computational burden on the tracking speed. Moreover, how to use natural language to guide the learning of graph-based structural feature representations remains not studied yet.

To handle above problems, we propose a novel structure-aware deep neural network that is end-to-end trained for visual tracking.  On the one hand, we utilize the graph convolutional network (GCN) to model the relations among training samples. Specifically, we first take them as graph nodes and use the standard convolutional network to extract their features. To fully utilize the spatial and temporal relations among samples (\emph{i.e.}, graph nodes), the deeply learned messages are then propagated among nodes via GCN to update and refine the pairwise relation feature for each node. After that, we form the final feature representation for each proposal by concatenating the enhanced and the original feature. On the other hand, we treat natural language embedding as high-level semantic information to guide the structural feature learning in the training phrase with triplet loss function, as shown in Figure \ref{differencewithMDNet} (b).

In the visual tracking task, the targets are easily lost when heavy occlusion occurs and they are out of view. It is difficult to re-track the targets when they are back to view as online update scheme adopted in most tracking methods will contaminate tracking models and the used local search strategy is also limited to recover the targets. Although some trackers employ the strategy of target re-detection~\cite{liu2018robust}, how to judge whether tracking failures occur or not is a challenging problem, and the re-detection models are too weak to recover the targets effectively when they reappear. To handle this problem, we elaborately design a novel subnetwork to learn the target-driven visual attentions from the guidance of both visual and natural language cues. Specifically, we use convolutional network to encode all the input data, i.e., the whole video frame, target object patch and natural language specification, for more efficient computation. The features are concatenated and input to an upsample module to generate the target-driven attention maps. The global proposals can be extracted from the attention regions and then input to the classifier together with local proposals. Therefore, in addition to providing complementary proposals to local ones, the global proposals could cover the targets well when they are back to view from heavy occlusion and out of view.

Generally speaking, our proposed algorithm is more intelligent by mining more structure information and exploring high-quality global proposal generation via target-driven visual attention. The contributions of this paper can be summarized as the following three aspects:  
	\begin{itemize}
	\item We propose an effective approach to handle the challenges of significant appearance changes, heavy occlusion and out of view in visual tracking. Extensive experiments on five tracking benchmarks against some recent and state-of-the-art trackers demonstrate that our proposed tracker is more robust to aforementioned challenging factors. 
	\item We propose a novel structure-aware deep neural network to make best use of the structures between training sample pairs and thus enhance the discriminative ability of feature representations. To make feature representations more discriminative, we introduce the natural language of target objects to assist visual feature learning via a triplet loss function.
	\item We elaborately design a novel global proposal generation network to the target-driven visual attentions from the guidance of both visual and natural language cues. Benefit from the global proposals, our tracker is able to re-track the target objects that are lost caused by the challenges of heavy occlusion and out of view.
	\end{itemize}

\begin{figure*}[t]
\center
\includegraphics[width=6in]{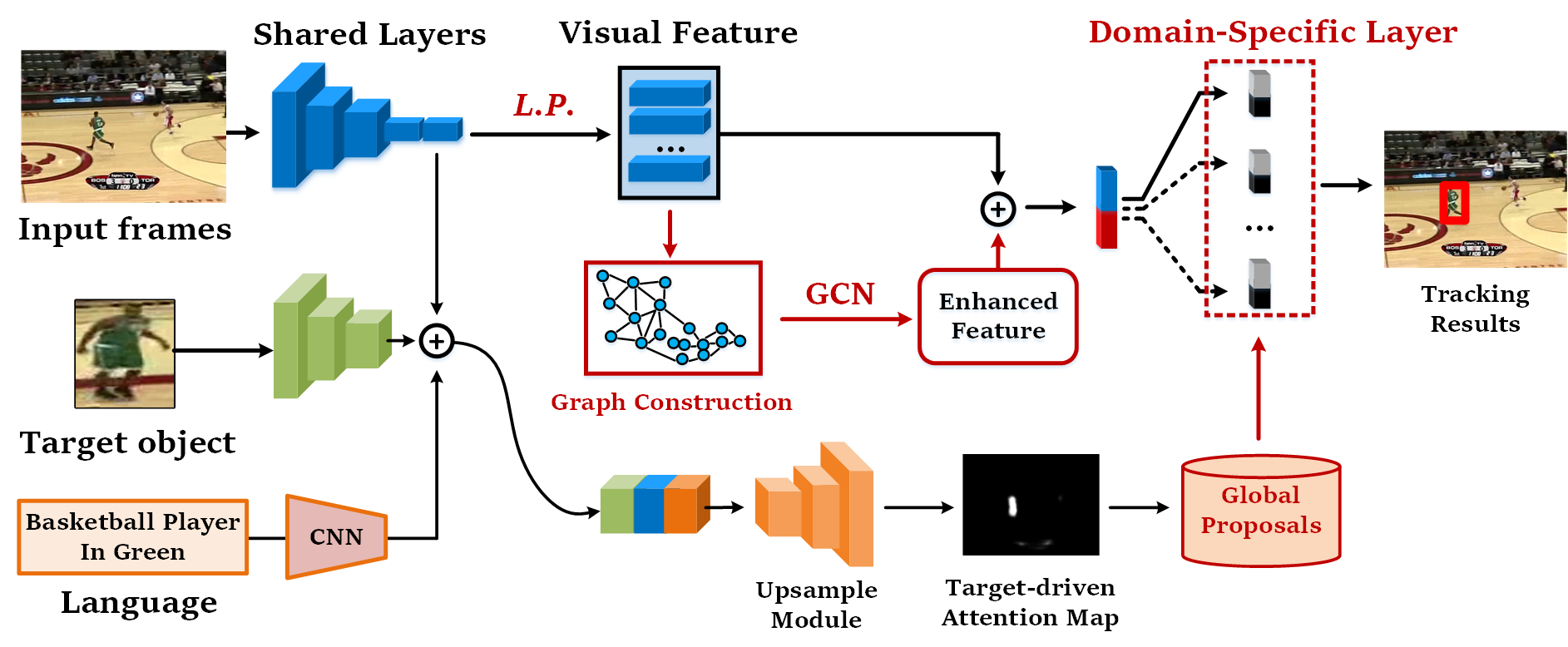}
\caption{The pipeline of our proposed tracking algorithm. }
\label{pipelineAll}
\end{figure*}

\section{Related Works} 
We give a brief review about tracking algorithms related to this paper as follows. 

\textbf{Structure based Trackers.} The algorithm to track non-rigid object has attracted great attention in recent years. Regular trackers can nearly handle the extreme deformations, therefore, some researchers begin to study this task and attempt to exploit part information and achieve promising performance. Son \emph{et al.} \cite{son2015tracking} utilize the online gradient boosting decision tree operation on individual patches to achieve robust visual tracking. Yeo \emph{et al.} in \cite{yeo2017superpixel} attempt to use Markov Chain on superpixel graph, however, the information propagation through a graph could be slow based on the structure. Ting and Yang \emph{et al.} propose the patch-based trackers based on correlation filter and combine the patches within a particle filter framework in \cite{li2015reliable} and \cite{liu2015real}, respectively. The major issue existed in these trackers is that they are all separately learn the correlation filter for each part and record the relative positions between each part and target center. Besides, these part-based trackers divide the target object into fixed number of fragments in a violent way. Hence, their model contains few discriminative local structure information due to such rough patch dividing, and little semantic information maintained in the feature of such patches. It is also very hard to design a reasonable updating strategy for these trackers and will be rather sensitive to model drift when drastic deformation occurred. Zhang \emph{et al.} \cite{Zhang_2018_ECCV} propose a structure constrained part-based model for visual tracking using DNNs and does not explicitly divide the target into parts. Their tracker can suppress the influence of aforementioned issues to some extent, however, they may still not be able to handle the drastic deformation or heavy occlusion well due to the lack of considering global information. Our model takes the target object and its language description as condition and estimate target-driven attention maps for global proposal generation which can handle this issue well.

\textbf{Multi-domain based Trackers.}
The idea of use multi-domain layers for the training of CNN is first proposed by Nam \emph{et al.} in \cite{Nam2015Learning}. They pretrain a CNN using a large set of videos with tracking groundtruths to obtain a generic target representation. Their network is composed of shared layers and multiple branches of domain-specific layers, where domains correspond to individual training sequences and each branch is responsible for binary classification to identify target in each domain. Their final tracking performance is indeed great and many trackers are developed based on this idea, such as BranchOut \cite{han2017branchout}, Meta-tracker \cite{Park_2018_ECCV}, Real-time MDNet \cite{Jung_2018_ECCV}. Although these trackers are all attempt to improve MDNet from different views, however, none of them consider the structure information when pretrain their models. In addition, these trackers still adopt the local search strategy which may make them sensitive to challenging factors as mentioned above. Our tracker utilizes the GCN and natural language to take the structure information into consideration and also joint use the global and local proposals for classification which make the baseline tracker more robust to challenging factors. 


\textbf{Visual Attention based Trackers.}
To handle the influence of video noises and/or tracker noises in the extremely challenging conditions, there are several attempts to combine attention maps with visual tracking. Choi \emph{et al.} \cite{Choi2016Visual} presented an attention-modulated visual tracking algorithm that decomposes an object into multiple cognitive units and trains multiple elementary trackers to modulate the distribution of attention based on various features and kernel types. Han \emph{et al.} \cite{Hong2015Online} proposed an online visual tracking algorithm by learning discriminative saliency map using CNN. They also directly searched the target object from attention locations. The spatial weights, which are widely used by DCF trackers to suppress the boundary effect, can also be interpreted as one type of visual attention. For example, the cosine window map \cite{bolme2010visual} and the Gaussian window map \cite{Danelljan2016Beyond} \cite{danelljan2015learning}. Recently, a number of efforts \cite{choi2017attentional} \cite{chu2017online} \cite{Yang_2018_ECCV} have been made to exploit visual attention within deep models. These approaches emphasize attentive features and resort to additional attention modules to generate feature weights. However, the feature weights learned in single frames are unlikely to enable classifiers to concentrate on robust features over a long temporal span. Moreover, slight inaccuracy of feature weights will exacerbate the misclassification problem. This requires an in-depth investigation on how to best exploit the visual attention of deep classifiers so that they can attend to target objects over time. Similar views can also be found in \cite{pu2018deep}. Instead, our proposed target-driven attention network takes video frames, initial target object and natural language as inputs. The generated attention maps are video-specific and can provide high-quality global proposals for visual tracking.

\textbf{Tracking by Natural Language.}
Integrating natural language into the computer vision community has becoming a new trend and many new tasks has been proposed, such as image caption, visual question answer, segmentation with natural language. The bridge used to connect the natural language and computer vision is the embedding technique which has achieved great progress in recent years, such as word2vector \cite{mikolov2013efficient}, GloVe \cite{Pennington2014Glove}. Usually, they utilize the memory network (RNN, LSTM, GRU or SRU) or CNNs to further learning the feature representations based on embedded vectors. They also integrate attention mechanism into their deep models to further improve the final performance. Improving tracking performance with natural language has been studied in \cite{Li2017Tracking}, they propose three kinds of models to fully illusturate possible combinations of visual tracking and natural language specifications. Different from their work, we embedding the natural language with CNN and use the embedding features to guide the global target-driven attention map generation. In addition, we also utilize the language embedding as high-level semantic information for shared feature learning.

\section{The Proposed Method}
The motivation of our method lies in two main aspects: i). How to learn a more robust deep feature representation by considering the correlations between extracted proposals? ii). How to obtain high-quality global proposals for visual tracking? In this paper, we propose an unified deep visual tracking algorithm guided by natural language specification, as shown in Figure \ref{pipelineAll}. We will give a detailed introduction to our tracker in following sections, including network architecture, loss functions, online tracking procedure and implementation details. 

\subsection{Network Architecture.} 
Our tracker contains two sub-networks, \emph{i.e.} structure-aware local search sub-network (SALNet) and global proposal generation sub-network (GPGNet). 

\subsubsection{SALNet}
The SALNet is actually a binary classification based visual tracker which follows the regular tracking-by-detection framework. Following MDNet, we use a deep convolutional network architecture as shown in Figure \ref{pipelineAll}. It takes a $107 \times 107$ RGB image patch as input, and contain five hidden layers including three convolutional layers and two fully connected layers. The convolutional layers are identical to the corresponding parts of VGG-M network \cite{chatfield2014return} \footnote{{https://github.com/HyeonseobNam/py-MDNet}} except that the feature map sizes are adjusted by our input size. The next two fully connected layers contain 4608 and 512 output units and are combined with ReLUs and dropouts. We adopt the feature from the second fc layer which is 512-D to denote corresponding image patch.  

The major difference between our SALNet with existing binary classification based trackers is that we take the correlations between training samples into consideration. Specifically, we formulate the deep feature learning problem for visual tracking as a \emph{node-focused} graph application. Given the features of extracted training samples, we can construct an undirected complete graph $G(V, E)$, where $V = \{v_1, v_2, ... , v_N\}$ denotes the set of nodes. Each node represents a feature vector of extracted image patch. We also establish edges $E$ on the graph $G$ to represent the set of relationships between different nodes. In this graph, we connect pairs of semantically related nodes together. More specifically, we will assign the weight based on the Euclidean distance between each paired proposal. We use $W_{ij}$ to denote the relation importance between node $i$ and node $j$, which can be represented as  following: 
\begin{equation}
\label{affinityFunction}
W_{ij}  =
\left\{
\begin{aligned}
\ &	 \frac{exp(S(g_i, g_j))}{\sum_{j} exp(S(g_i, g_j))},    ~~~ if ~~~ i \neq j \\
 &	0,						~~~~~~~~~~~~~~~~~~~~~~~~~~~~~~~ else \\
\end{aligned}
\right.
\end{equation}
where $g_i$ and $g_j$ are the $i$-th and $j$-th node. $S()$ is a pairwise similarity estimation function, that estimates the similarity score between $g_i$ and $g_j$.

After the affinity matrix is computed according to Eq. \ref{affinityFunction}, we perform normalization on each row of the matrix so that the sum of all the edge values connected to one proposal $i$ will be 1. Following \cite{vaswani2017attention}, we adopt the softmax function for the normalization: 
\begin{equation} 
\label{softmaxFunction}
G_{ij} = \frac{exp(W_{ij})}{\sum_{j=1}^{N}exp(W_{ij})}
\end{equation}
The normalized $G$ is taken as the adjacency matrix representing the similarity graph. 

Different from standard convolution that operate on local region in an image, the convolutional operations on graphs is then defined by computing the response at a node based on the neighboring nodes defined by the adjacency graph. Mathematically, the convolutional operations for each layer in the network is represented as: 
\begin{equation}
\label{graphConvolution}
Z = \hat{A} X' W 
\end{equation}	
where $\hat{A}$ is a normalized version of the binary adjacency matrix $A$ of the graph, with $n \times n$ dimensions. $X'$ is the input $n \times k$ feature matrix from previous layer. $W$ is the weight matrix of the layer with dimension $k \times c$, where $c$ is the output channel number. Therefore, the input to a convolutional layer is $n \times k$, and the output is a $n \times c$ matrix $Z$. The convolution operations can  be stacked one after another. A non-linear operation (ReLU) can be applied after each convolutional layer. For the final convolutional layer, the number of output channels is the number of label classes ($c = C$) in the supervised learning. In this paper, we only want to obtain its output feature, therefore, we do not integrate this layer into our network. More detailed introductions can be found in \cite{2016Semigcn} \cite{battaglia2018relational}.  

After we obtain the enhanced feature via GCN, we concatenate it with original input as the final feature representation of each proposal. Following MDNet, we also introduce the domain-specific layers to model the correlations between different video sequences in the training dataset. We prefer readers to check the MDNet to have a deeper understand of this algorithm.

In the shared feature learning phase (\emph{i.e.} the SALNet), the loss function used for binary classification (\emph{i.e.} BCE loss) can be formulated as: 
\begin{equation}
\label{crossEntropyLoss} 
L_c = - \sum_{i=1}^{T} y_i log p_i  + (1-y_i) log (1-p_i)
\end{equation}
where $T$ is the mini-batch size, $y_i$ the ground truth label of the $i$-th sample, $p_i$ is the prediction of corresponding sample from deep neural network.

In addition to BCE, the triplet loss function is also adopted to ensure that all positive samples $\textbf{V}^p$ (\emph{positive}) are closer to the high-level semantic vectors $\textbf{V}$ (\emph{anchor}) and all negative samples $\textbf{V}^n$ (negative) are at a distance from the anchor vector, as illustrated in Figure \ref{differencewithMDNet} (b). Formally, we have: 
\begin{equation}
\label{tripletLossfunction}
||V - V^p||_2^2 + \alpha \leq ||V - V^n||_2^2, ~~ \forall (V, V^p, V^n) \in \mathcal{T} 
\end{equation}
where $\alpha$ is a margin that is enforced between positive and negative pairs, we set it as 1.0 in our implementation. $\mathcal{T}$ is the set of all possible triplets in the training set and the mini-batch can be setted as N. Hence, the loss function for the mini-batch can be formulated as: 
\begin{equation}
\label{tripletLossfunction}
L_t = \sum_{i=1}^{T} [ ||V - V^p||_2^2 -  ||V - V^n||_2^2 + \alpha ]_+  
\end{equation}

Therefore, the final loss function for the optimization of the SALNet can be formulated as: 
\begin{equation}
\label{totalLossfunction}
Loss = L_c + \lambda * L_t 
\end{equation}
where $\lambda$ is a tradeoff parameter, we experimentally set it as 0.1 in our experiments. 

\subsubsection{GPGNet}
Although the proposed SALNet already achieve good performance on some video sequences, however, it still cannot get rid of the issues caused by local search strategy under the tracking-by-detection framework. In this paper, we propose the global proposal generation network (GPGNet) to complement with local proposals for robust visual tracking. 

\begin{figure}[t]
\center
\includegraphics[width=3.3in]{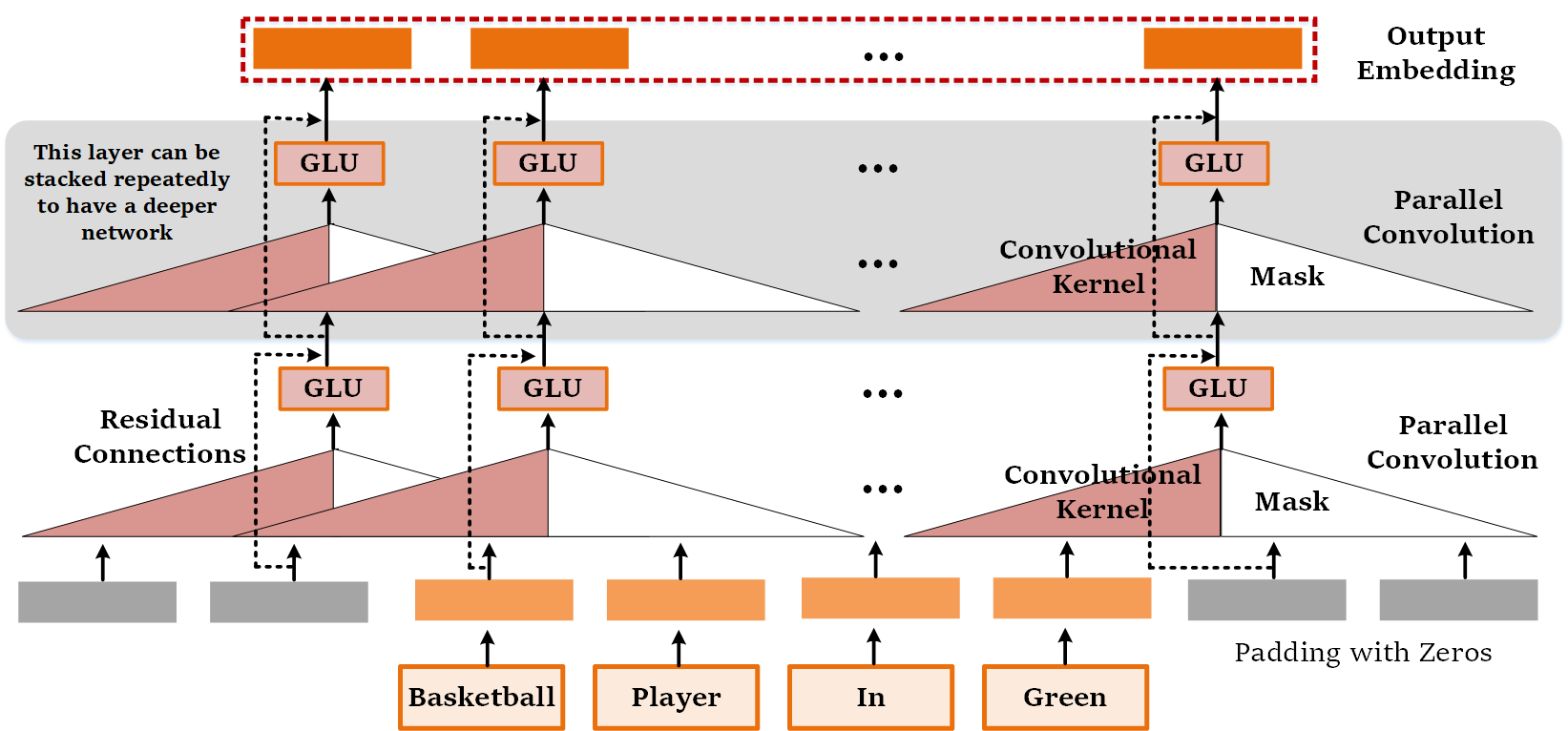}
\caption{The illustration of convolutional network for natural language embedding. }
\label{cnnLanEmbedding}
\end{figure} 

As shown in Figure \ref{pipelineAll}, the inputs of this module are target object patch, video frame and natural language specification. For all the input data, we only use convolutional networks to obtain its features due to the efficiency of CNN. Specifically, for each video frame and the target object patch, we resize them into $192 \times 256 \times 3$ and obtain their feature map whose dimension is $12 \times 16 \times 512$ via VGG-Net. For the natural language, we first embedding each word into a 512-D feature vector. Different from regular operation which use RNN model to embedding the language \cite{Li2017Tracking} \cite{Margffoy_Tuay_2018_ECCV} \cite{Chen_2018_ECCV}, we adopt CNN to encode them with more efficient parallel convolution which has widely used in many tasks \cite{gehring2017convolutional} \cite{Aneja_2018_CVPR}. The detailed configuration of the convolutional network can be found in Figure \ref{cnnLanEmbedding}. The visual features of video frame and target object are concatenated along the channel, hence, we obtained a feature map whose dimension is $12 \times 16 \times 1024$. The maximum length of given sentence is 16 in this paper, therefore, we can obtain the sentence embedding  $16 \times 512$. Then, we expand this embedding into $12 \times 16 \times 512$ and obtain a feature map $12 \times 16 \times 1536$ after concatenate with visual features. The features we obtained in the encoding phase are concatenated together and input into the upsample network (which is a reversed VGG-Net). 

Following \cite{Margffoy_Tuay_2018_ECCV}, we adopt the binary cross-entropy loss for the optimization of GPGNet.

\subsection{Online Tracking}
When tracking the target object in a new sequence with the guide of target-driven attention maps, the shared layers in pre-trained CNN and a new binary classification layer are combined together to construct a new network. And also the GCN module  is only used for shared feature learning in the training phrase to achieve more efficient tracking. Online tracking is performed by evaluating the candidate windows randomly sampled around the previous target state and proposals extracted from attention regions. In this paper, to estimate the target state in each frame, $N$ target candidates $\textbf{x}^1, \textbf{x}^2, ... , \textbf{x}^N$  sampled around the previous target state and current attention regions are evaluated using the network. We obtain their positive scores $F^+(\textbf{x}^i)$ and negative scores $F^-(\textbf{x}^i)$ from the network. During the sampling phase, our target-driven attention maps could provide proposals with more accurate location and scale information, which could make the searching process more efficient and effective. Hence, the optimal target state $\textbf{x}^*$ is given by finding the example with the maximum positive score as:
\begin{equation}
\label{MDNETMaximum}
\textbf{x}^* = \arg \max_{\textbf{x}^i} F^+(\textbf{x}^i)
\end{equation}	

We adopt the same strategy with MDNet, \emph{i.e.}, long-term and short-term updates, to update our model. Long-term updates are performed in regular intervals using the positive samples collected for a long period while short-term updates are conducted whenever potential tracking failures are detected (when the positive score of the estimated target is less than 0.5) using the positive samples in a short-term period.

\subsection{Implementation Details}
The global proposals generated from attention maps can be concluded as the following three steps: 1) Obtain attention regions and center location of each region, given the attention map; 2) Obtain BBox, which attempts to cover each attention region; 3) Employ Gaussian sampling strategy on these bounding boxes to generate proposals. 

\begin{figure}[htb]
\center
\includegraphics[width=3.3in]{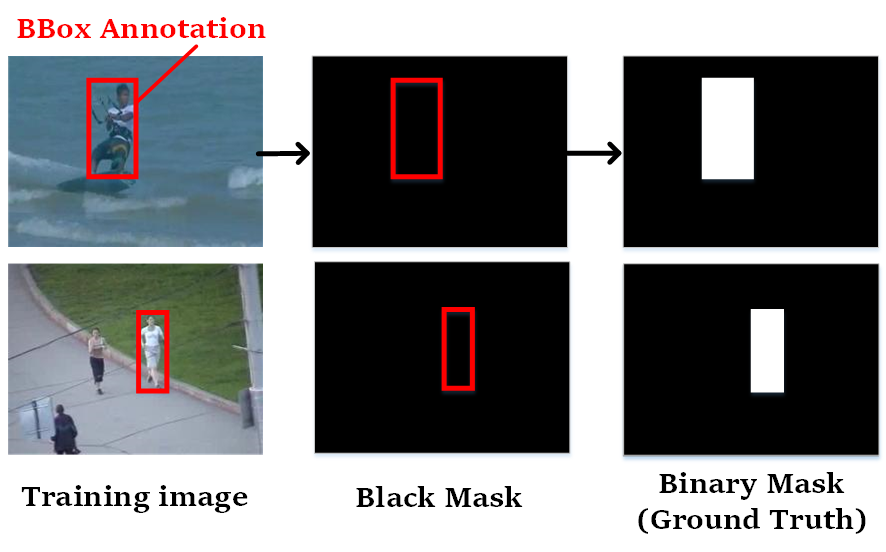}
\caption{The pipeline of pre-processing training samples for GPGNet.}
\label{GPGtrainSamples}
\end{figure}

The GPGNet is used to generate the target-driven attention maps for global proposal extractation. We use binary mask for the training of this network which can be obtained directly from BBox annotations. As shown in Figure \ref{GPGtrainSamples}, we first generate a black mask which has the same resolution as the video frame, then, we white the target object regions according to annotated BBox in the training dataset. The binary mask is used as the ground truth attention maps to optimize the GPGNet. Following the regular semantic segmentation,  we adopt the binary cross-entropy loss to measure the difference between the generated attention maps and the ground truth. 

The training details of SALNet: initial learning rate is 0.0001; batch size is set as 8, and Adam is utilized for optimization. For the GPGNet, the initial learning rate is 5e-5, batch size is 20, Adagrad is used for the optimization. Three convolutional layers are used to encode the natural language and the maximum sentence length is 16. We train this network for 50 epochs. All the experiments are implemented based on PyTorch on a desktop computer with Ubuntu 16.04, I7-6700k, NVIDIA TITAN Xp with 12G VRAM and 32G RAM.

 
\section{Experiments}

We will first introduce the dataset and evaluation criteria used in this paper. After that, we will compare our tracking results with other state-of-the-art visual trackers on several public benchmarks. Then, we conduct ablation studies to validate the effectiveness of each component. Finally, we will discuss the difference with existing trackers. 
 
\subsection{Datasets and Evaluation Criterion}
The training datasets used in this paper for SALNet and GPGNet are \textbf{TLP50} \cite{moudgil2017long}, \textbf{DTB70} \cite{li2017visualDLP70} and \textbf{LaSOT} \cite{fan2018lasot} \footnote{{https://cis.temple.edu/lasot/}} dataset which totally contain 120 (50 + 70) and 1400 video sequences, respectively \footnote{The baseline method pyMDNet used in this paper is implemented based on PyTorch and pre-trained on two long-term dataset \textbf{TLP} \cite{moudgil2017long} and \textbf{DTB70} dataset \cite{li2017visualDLP70} for all our experiments.}.  The LaSOT provide both the BBox and natural language annotations of target object, which is suitable for our natural language guided tracking task. Specifically, we generate binary mask for each video frame by setting the target object pixels as zero and background pixels as 255. We use those masks as goundtruth attention maps to optimize the GPGNet. It is also worthy to note that, we only select 44660 images from the LaSOT dataset (it totally contains 3.52 million frames) for the training of GPGNet to quickly validate our proposed method.  The testing is conducted on five benchmark datasets, including \textbf{OTB-2013} \cite{WuLimYang13}, \textbf{OTB-100} \cite{wu2015object}, \textbf{VOT-2014} \cite{Kristan2014a},  \textbf{VOT-2016} \cite{VOT_TPAMI} and \textbf{TC128} \cite{Liang2015Encoding} dataset. The natural language specification of OTB100 is borrowed from lang-tracker \footnote{{https://github.com/QUVA-Lab/lang-tracker}}, and other datasets used for testing is annotated by one person to maintain its consistency.

Two widely used evaluation protocols are utilized in this paper: \textbf{{success rate}} and \textbf{{precision rate}}. These two criteria are all aiming at measure the percentage of successfully tracked frames. For the success rate, a frame is declared to be successfully tracked if the estimated bounding box and the ground truth box have an interaction-over-union overlap larger than a certain threshold. For precision rate, tracking on a frame is considered successful if the distance between the center of the predicted box and the ground truth box is under some threshold.

\begin{figure*}[t]
\center
\includegraphics[width=7in]{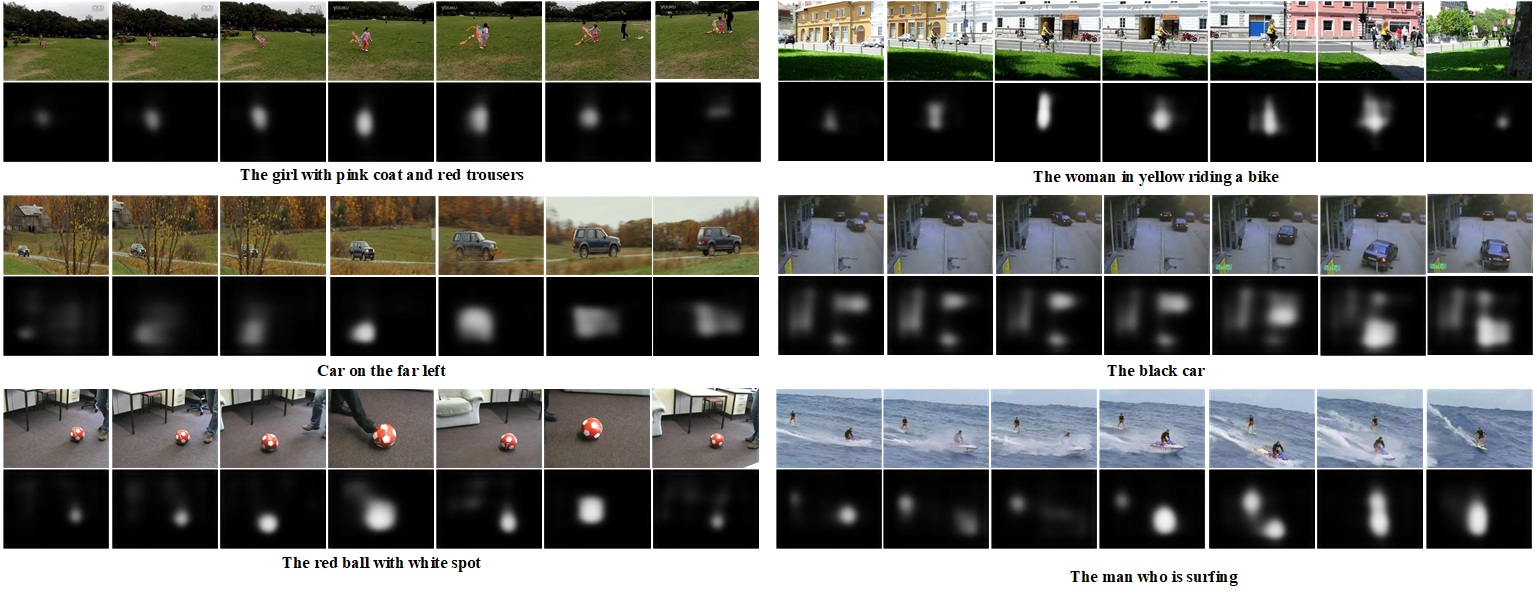}
\caption{The attention maps generated by our GPGNet.}
\label{attentionShown}
\end{figure*}

\begin{figure*}[t]
\center
\includegraphics[width=7in]{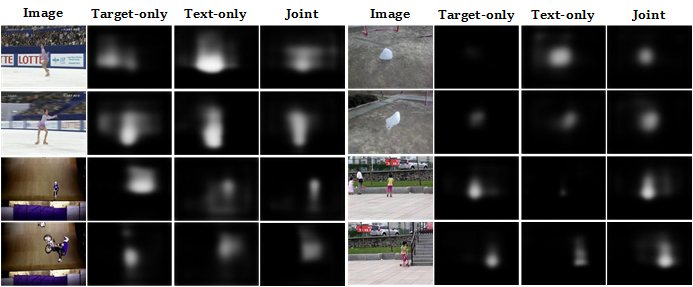}
\caption{The difference between target object only, natural language only and joint based global attention estimation.}
\label{attentionShown-v-t-joint}
\end{figure*}

\begin{table*}[htp!]
\center
\scriptsize
\caption{The tracking results on OTB-2013 and OTB-100 Benchmark (The top three results are highlighted in \textbf{{\color{red} red}}, \textbf{{\color{green} green}} and \textbf{{\color{blue} blue}}, respectively).} \label{OTBbenchmarkresults}
\begin{tabular}{l|c|c|c|c|c|c|c|c|c}
\hline
\hline
\textbf{Algorithm}				&\textbf{SINT++}      &\textbf{StructSiam} 		&\textbf{CREST}   		&\textbf{CCOT} 		  &\textbf{RFL  } 			&\textbf{RASNet}	 	 &\textbf{SRDCFD}	 	&\textbf{CSR-DCF} &\textbf{ AFCN}  \\
\hline 
\textbf{OTB-2013}  &0.839/0.624	&0.880/0.638 &\textbf{{\color{green} 0.908}}/\textbf{{\color{green} 0.673}}		&0.899/ \textbf{{\color{blue} 0.672}}   &0.786/0.583    &0.892/0.670  &0.870/0.653  	&-/- 			  & 0.860/0.607	 \\
\hline
\textbf{OTB-100} 	&0.768/0.574 &0.851/0.621 &0.838/0.623 	&\textbf{{\color{red} 0.898}}/\textbf{{\color{red} 0.671}}  &0.778/0.581    &-/0.641 			&0.825/0.627  		&0.733/0.587  	  &0.799/0.573  	\\
\hline 
\hline 
\textbf{Algorithm}				&\textbf{ADNet} 		&\textbf{Meta-Tracker}		&\textbf{MemTrack}  &\textbf{Staple}  &\textbf{CFNet} &\textbf{Lang-Tracker}&\textbf{SiamFC}&\textbf{pyMDNet}&\textbf{Our}\\ 
\hline 
\textbf{OTB-2013}	&\textbf{{\color{blue} 0.903}}/0.659	 	&-/-			&0.849/0.642 		&0.793/0.600	 	&0.785/0.589			&-/0.578   		&0.809/0.607	 	&0.880/0.655			&\textbf{{\color{red} 0.925}}/ \textbf{{\color{red} 0.676}}      \\ 
\hline
\textbf{OTB-100}	&\textbf{{\color{blue} 0.880}}/\textbf{{\color{green} 0.646}}	 	&0.856/0.637	 &0.820/0.626  		&0.784/0.581	 	&0.777/0.586			&-/-   		&0.771/0.582	 	&0.866/\textbf{{\color{blue} 0.643}}			&\textbf{{\color{green} 0.889}}/\textbf{{\color{green} 0.646}}         \\ 
\hline 
\end{tabular}
\end{table*}

\begin{figure*}[t]
\center
\includegraphics[width=6.5in]{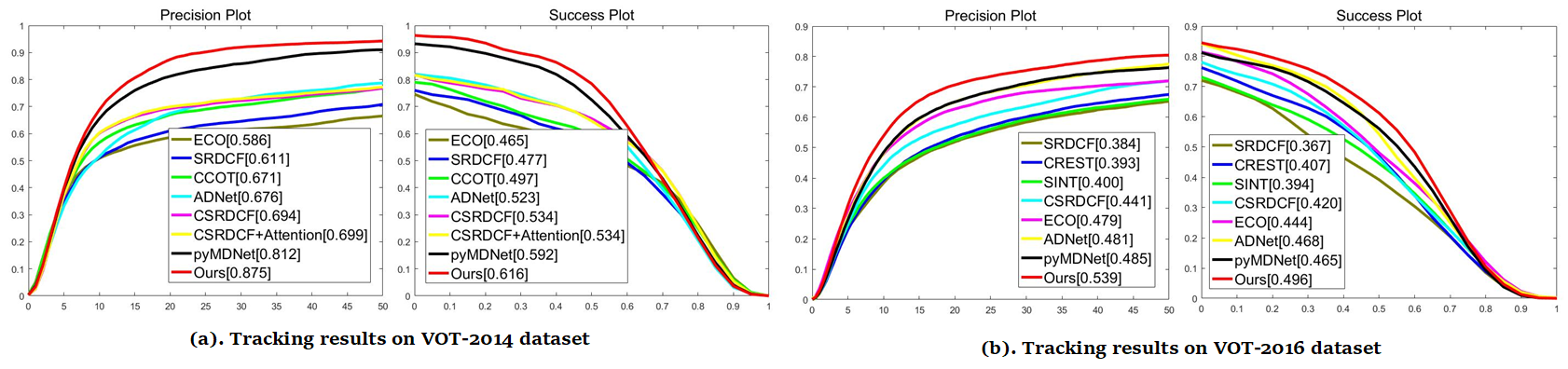}
\caption{The tracking results on VOT-2014 (left two sub-figures) and VOT-2016 dataset (right two sub-figures). }
\label{resultsVOT}
\end{figure*}	

\begin{figure}[t]
\center
\includegraphics[width=3.3in]{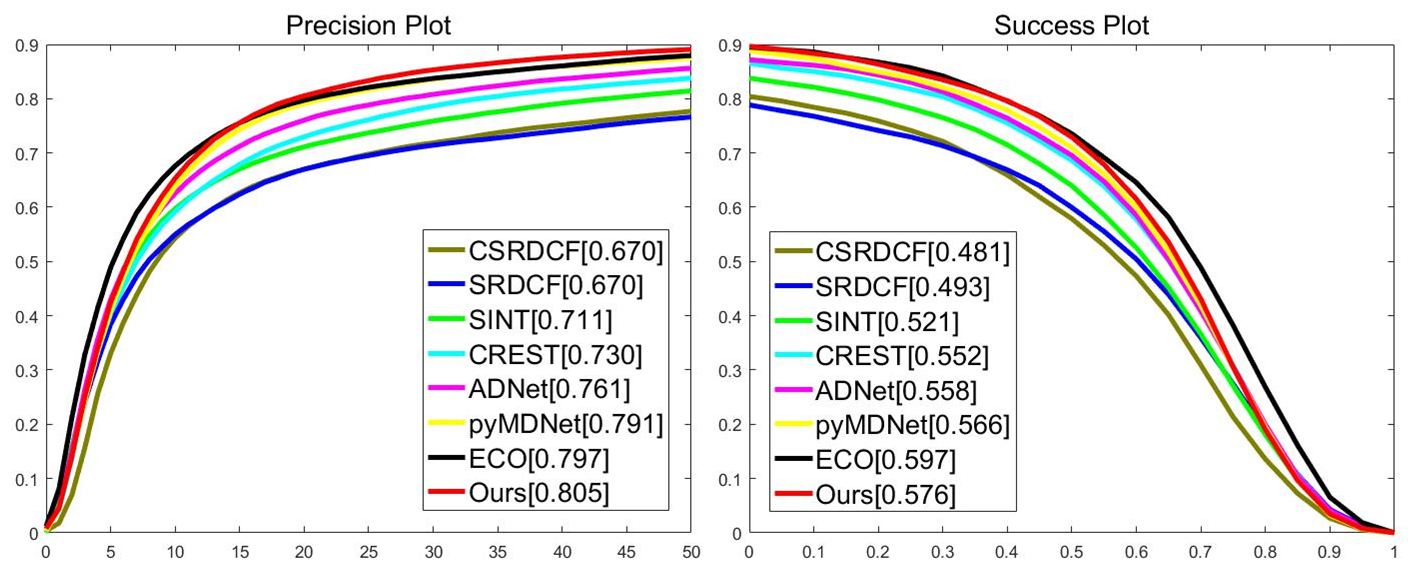}
\caption{The tracking results on TC128 dataset. }
\label{tcresults}
\end{figure}

\subsection{Comparison with State-of-the-art Trackers}
To fully demonstrate the effectiveness of our proposed algorithm, we compare with many recent and popular trackers, including: \textbf{MDNet} \cite{Nam2015Learning}, \textbf{CCOT} \cite{Danelljan2016Beyond}, \textbf{ECO} \cite{Danelljan2017ECO}, \textbf{RASNet} \cite{Wang_2018_CVPR}, \textbf{SINT} \cite{Tao2016Siamese}, \textbf{SINT++} \cite{Wang_2018_CVPR}, \textbf{CSR-DCF} \cite{Lukezic2016Discriminative}, \textbf{ADNet} \cite{Yun2017Action}, \textbf{Meta-Tracker} \cite{Park_2018_ECCV}, \textbf{MemTrack} \cite{Yang_2018_ECCV}, \textbf{Staple} \cite{bertinetto2016staple}, \textbf{CFNet} \cite{valmadre2017end}, \textbf{Lang-Tracker}  \cite{Li2017Tracking}, \textbf{SiamFC} \cite{bertinetto2016fully}, \textbf{ReGLe} \cite{li2017regle}, \textbf{StructSiam} \cite{Zhang_2018_ECCV} and \textbf{AFCN} \cite{choi2017attentional}.

As shown in Table \ref{OTBbenchmarkresults}, it is obvious to find that our proposed visual tracker achieve good and even better performance than some recent trackers on OTB-2013 and OTB-100 benchmark datasets. Compared with the baseline method pyMDNet, our method improves significantly on both OTB-2013 and OTB-100 dataset. Specifically, our algorithm improves the tracking accuracy (precision plot/success plot) from 0.880/0.655 to 0.925/0.676 on OTB-2013 dataset; from 0.866/0.643 to 0.889/0.646 on OTB-2015 dataset, respectively. Our method also achieves good tracking results on the public benchmarks compared with other visual trackers. Our tracker does not perform as well as the top performing tracker CCOT on OTB100 dataset. It is because: i).  CCOT crops the sample in a continuous space for scale estimation, while our tracker only randomly draws a sparse set of samples. ii). CCOT also use multiple features (e.g., color names, HOG and deep features), while we only use deep features. We will consider to explore these ideas as our future works. 

\begin{figure*}[t]
\center
\includegraphics[width=7in]{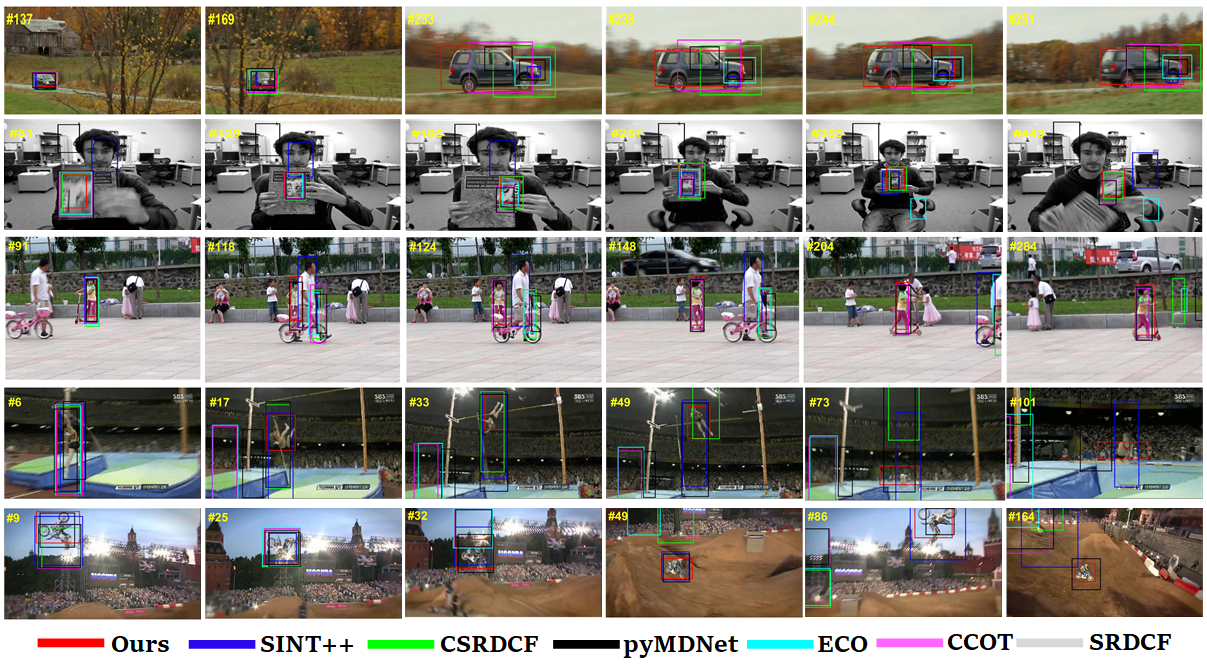}
\caption{The tracking results of our method and other trackers.}
\label{trackingResultsFigure}
\end{figure*}

We also show the tracking results on VOT-2016 which are evaluated with its own default metrics, as shown in Table \ref{VOTOriResults}. We can find that our tracker can achieve comparable or even better performance when compared with other trackers. The illustration about the tracking results and target-driven attention maps can be found from Figure \ref{attentionShown}, \ref{attentionShown-v-t-joint} and \ref{trackingResultsFigure}. 

\begin{table*}[htp!]
\small 
\center
\caption{ Comparison with other trackers on VOT-2016 dataset with default metrics.} \label{VOTOriResults}
\begin{tabular}{c|ccccccccc}
\hline
\hline
{Algorithm}	 &{CCOT }   &{EBT }		 &{Staple }		&{SRDCF }	&{HCF } &{SiamRN }	&{DSST}   	 &{MDNet}  &{Ours}\\
\hline
{EAO}  &0.3310   &0.2913     &0.2952   &0.2471   &0.2203   &0.2766   &0.1814     &0.2572   &0.3045   \\
\hline 
{FPS}  &82.18   &2.87   &14.43  &503.18   &328.73      &7.05   &13.90    &2.66   &2.27   \\
\hline 
\end{tabular}
\end{table*}

\begin{table}[htp!]
\center
\scriptsize
\caption{ Tracking performance without or with global proposals on OTB100 dataset. The tracking results on precision plot and success plot are listed as follows.} \label{GPResults}
\begin{tabular}{l|c|c|c|c}
\hline
\hline
\textbf{Algorithm}	 &\textbf{Our-SALNet}   &\textbf{Our-TO}	 &\textbf{Our-NL}	 &\textbf{Our-JTNL}\\
\hline
\textbf{OTB100}  &0.876/0.644  &0.886/0.647  &0.884/0.643  &0.889/0.646\\
\hline
\end{tabular}
\end{table}

\subsection{Ablation Studies}

\textbf{The Effect of GCN.} 
To demonstrate the effectiveness of structured information from other proposals, we conduct experiment on this component (\emph{i.e.} pyMDNet+GCN) on OTB-2013 dataset. As shown in Table  \ref{CAResults}, pyMDNet+GCN improved the tracking result from 0.880/0.655 to 0.905/0.671 on precision and success plot, respectively, compared with baseline method pyMDNet. This result fully validated the effectiveness of the structured information from other nodes, that is to say, the GCN can help to learn more discriminative deep features for visual tracking.

\textbf{The Effect of Triplet Loss.} 
To validate the effectiveness of natural language guided feature learning, we test the model which only with triplet module, \emph{i.e.} pyMDNet+Language, as we can see from Table  \ref{CAResults}, the prior knowledge also improved the feature learning. This experiment fully demonstrates the effectiveness of the introduced natural language specification to guide the shared feature learning. 

\begin{table}[htp!]
\center
\scriptsize
\caption{ Component Analysis on OTB-2013 dataset. The tracking results on precision plot and success plot are listed as follows.} \label{CAResults}
\begin{tabular}{l|c|c|c}
\hline
\hline
{Algorithm}	 &\textbf{pyMDNet}   &\textbf{+ GCN}	 &\textbf{+ Language}	 \\
\hline
{OTB-2013}  &0.880/0.655  &0.905/0.671  &0.913/0.668  \\
\hline
\end{tabular}
\end{table}

\textbf{The Effect of Global Search Strategy.} 
As shown in Table \ref{GPResults},  we conduct object tracking without global proposals (Our-SALNet), and also joint use local and global search strategy for robust visual tracking. Specifically, we estimate the attention maps with following three versions: target object patch based (Our-TO), natural language based (Our-NL) and joint target and language (Our-JTNL) based. It is easy to find that the utilization of global search strategy can significantly improve the tracking results compared with baseline method Our-SALNet. We also visualize some of these global attention maps in Figure \ref{attentionShown}.

\textbf{The Generic of Target-driven Attention Maps.} 
We show our target-driven attention maps can also be integrated with other trackers, such as CSRDCF \cite{Lukezic_IJCV2018}. We take the attention maps generated by our GPGNet as a kind of feature representation, and integrated with CSRDCF for robust visual tracking. For example, CSR-DCF uses [gray, color name and hog feature] as original features. After integrated with our attention maps, its feature tuple becomes [gray, color name, hog, attention map].  As shown in Figure \ref{resultsVOT} (a), the tracking results of CSRDCF can be improved with our attention maps on VOT dataset.

\begin{figure}[t]
\center
\includegraphics[width=3.3in]{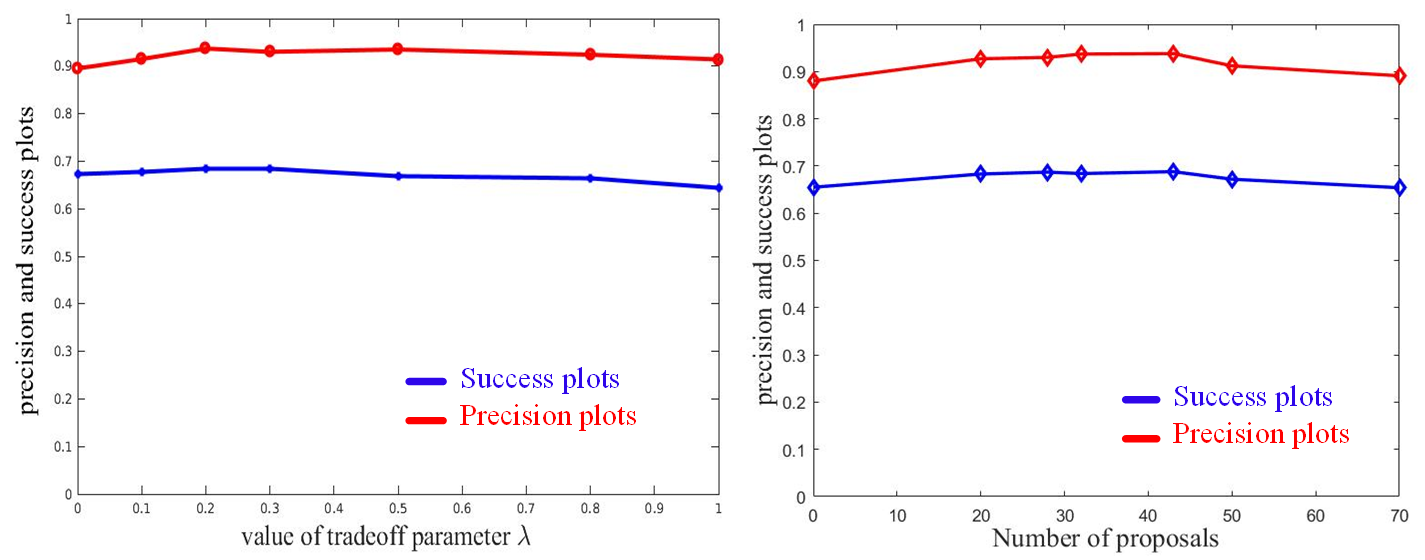}
\caption{The analysis of tradeoff parameter (left sub-figure) $\lambda$ and node numbers (right sub-figure). The {{\color{red} red}} line denotes the variation of precision plots, and the {{\color{blue} blue}} line denotes the success plots. }
\label{resultsAblationStudyNodeParameter}
\end{figure}

\textbf{Influence of Tradeoff Parameter $\lambda$.}
As shown in Eq. \ref{totalLossfunction}, our loss function contains a hyperparameter $\lambda$ which is introduced to tradeoff the classification loss and triplet loss. In this section, we conduct some experiments on this parameter (we set the $\lambda$ equal to 0, 0.1, 0.2, 0.3, 0.5, 0.8 and 1) to show its influence on the final tracking results. The curve to shown variation of tracking performance can be found in Figure \ref{resultsAblationStudyNodeParameter}.  We can find that our proposed deep model is not sensitive to the hyperparameter $\lambda$. 

\textbf{Influence of Node Numbers for GCN.}
To test the influence of different numbers of training samples, we extract different amount of samples (\emph{i.e.} 0,  20,   28,   32,   43,   50, 70 ) and conduct experiment to check the final result. As shown in Figure \ref{resultsAblationStudyNodeParameter}, the tracking results can be enhanced significantly when integrating GCN, since the result of other proposals are all better than zero's (i.e. the baseline method pyMDNet). We can also find that the results are better than others when the node number belong to the range of (30, 50).

\textbf{Influence on Different Layers of GCN.} 
To check the influence of different graph convolutional layers, we conduct ablation studies on this question. We set the number of GCN layers as 2, 3, 5 and 8 layers to pretrain the model and test on the OTB2013 dataset. As shown in Table \ref{differGCNResults}, the tracking results can be enhanced when increasing the number of GCN layers. However, it also increased the training time when more layers are added. Hence, we choose 3 GCN layers to achieve better tradeoff between accuracy and training time in our experiments.  

\begin{table}[htp!]
\center
\small 
\caption{ Tracking results with different layers of GCN on OTB2013 dataset.} \label{differGCNResults}
\begin{tabular}{l|ccccccc}
\hline
\hline
{Layers}	 &{2}	 &{3}	 &{5} &{8}\\
\hline
{SR}  &0.654  &0.663	&0.671 &0.670	\\
\hline
\end{tabular}
\end{table}

\begin{figure}[t]
\center
\includegraphics[width=3.3in]{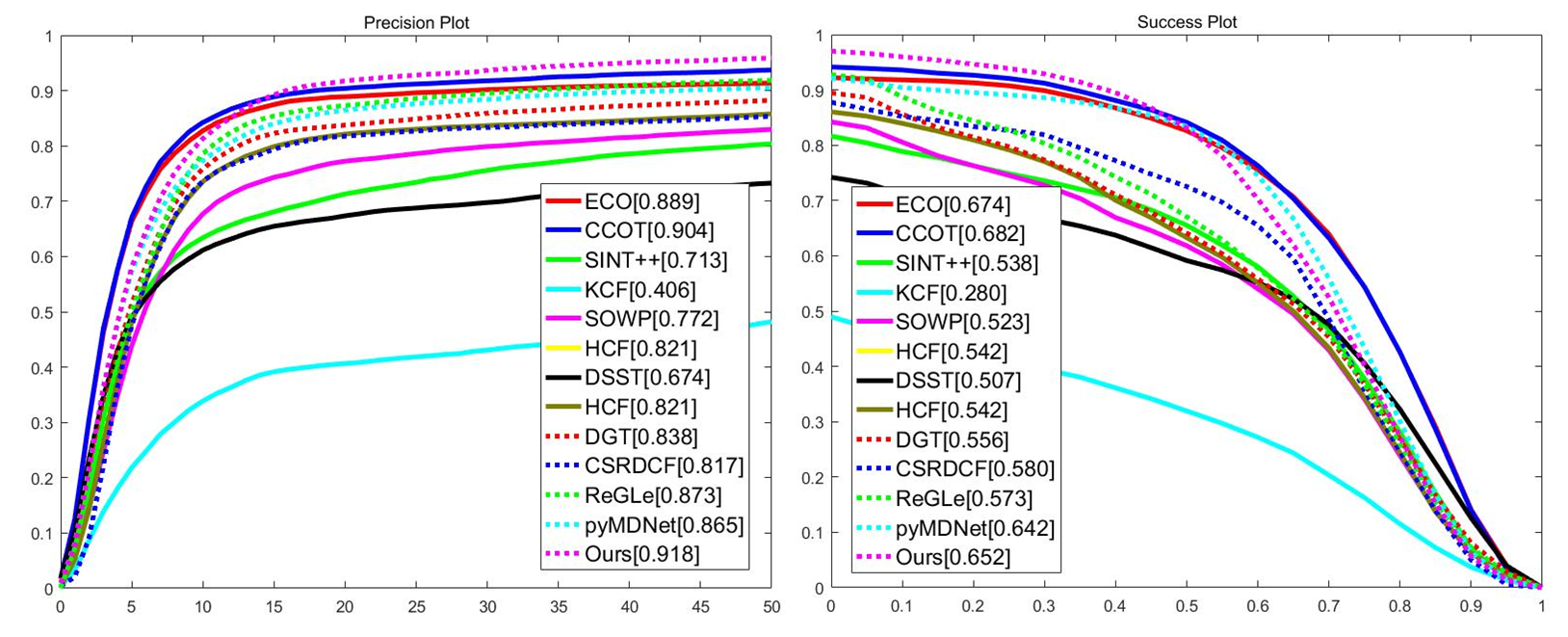}
\caption{The tracking results on 46 videos selected from OTB100 dataset (these videos all contain similar appearance objects with the target object).}
\label{similarAppearanceResults}
\end{figure}

\textbf{Tracking results on similar appearance videos.}
To validate the performance on videos with similar target objects, we also test our tracker on 46 video sequences \footnote{The selected video list: {Basketball, Bird1, Girl2, BlurCar1, BlurCar2, BlurCar4, Bolt, Bolt2, Walking, Walking2, BlurCar3, Freeman3, Car1, Car2, Car24, Car4, CarDark, Couple, Coupon, Crossing, Crowds, Deer, Football, Football1, Human3, Human4, Human5, Human6, Human7, Human8, Human9, Ironman, Jogging-1, Jogging-2, Jumping, KiteSurf, Liquor, Shaking, Singer1, Singer2, Skating1, Skating2-1, Skating2-2, Soccer, Subway, Suv.}} selected from OTB100 dataset. These videos contain at least one or more similar objects with target object. We want to validate the robustness of our target-driven attention maps via this experiments. As shown in Figure \ref{similarAppearanceResults}, our tracker can still achieve good performance on these challenging videos. Specifically, we can achieve 91.8/65.2 on this sub-dataset evaluated with PR and SR evaluation criterion. It is easy to find that our results are better than baseline method pyMDNet (86.5/64.2) and some other recent visual trackers, such as SINT++ \cite{Wang_2018_CVPR}, ReGLe \cite{li2017regle}. 

\subsection{Discussion}

\textbf{Difference Between Regular Saliency Estimation and Our Attention Maps.} Saliency maps usually focus on the target we humans attend, however, it maybe not the target we want to track in practical videos. Therefore, it can not be directly utilized in practical tracking algorithms. Meanwhile, our attention maps are generated based on initial target object and natural language specifications. It only focuses on the target object we want to track in each video, in another word, our attention maps are \emph{video-specific}.

\textbf{Difference with Existing Trackers.}
	The most relevant works with ours are Lang-Tracker \cite{Li2017Tracking} and MDNet. \textbf{For the Lang-Tracker}: lang-tracker use language to detect target object in the first frame and tracking target object according to image patch and language descriptions for subsequent frames; we use the natural languge for shared feature learning and global attention estimation, which also improve the final tracking results significantly. \textbf{For the MDNet}: i). MDNet did not consider the structure information between training samples or language; we model this information with GCN and triplet loss function when design our network. ii). MDNet only adopt the local search strategy by following tracking-by-detection framework which make their tracker rather sensitive to challenging factors; Our tracker jointly use local and global search strategy for robust visual tracking. Extensive experiments on five tracking benchmarks validated the effectiveness of our proposed algorithm.

\section{Conclusion}
In this paper, we propose a novel visual tracker, named DAT, to track the target object based on the provided BBox and its natural language specification. Our tracker can be devided into two main subnetworks: SALNet and GPGNet. The SALNet is a novel structure-aware deep neural network by take both the correlations between video sequences and training samples in each video into consideration. We adopt the softmax and triplet loss functions to train this sub-network. We also propose the GPGNet, which is a novel target-driven global attention estimation network to ensure the locations we should focus on. These proposals extracted from the attention regions are feed into binary classifier together with proposals extracted from local search window. The proposal with maximum response score will be chosen as the tracking result of current frame. Extensive experiments on several public tracking benchmarks validated the effectiveness of our proposed method.

{
\bibliographystyle{ieee}
\bibliography{reference}
}

\end{document}